# Analysis and Fully Memristor-based Reservoir Computing for Temporal Data Classification

*Ankur Singh[1], Sanghyeon Choi[2], Gunuk Wang[3], Maryaradhiya Daimari[1], and Byung-Geun Lee[1]**

[1]School of Electrical Engineering and Computer Science, Gwangju Institute of Science and Technology, Gwangju 61005, Republic of Korea (e-mail: ankursingh@gm.gist.ac.kr; maryaradhiya@gm.gist.ac.kr, bglee@gist.ac.kr).

[2]Department of Electrical and Computer Engineering, University of California, Santa Barbara, CA, 93106, USA (e-mail: chl299@ucsb.edu).

[3]KU-KIST Graduate School of Converging Science and Technology, and Department of Integrative Energy Engineering, Korea University, 145 Anam-ro, Seongbuk-gu, Seoul 02841, Republic of Korea (e-mail: gunukwang@korea.ac.kr).

**Abstract:** Reservoir computing (RC) offers a neuromorphic framework that is particularly effective for processing spatiotemporal signals. Known for its temporal processing prowess, RC significantly lowers training costs compared to conventional recurrent neural networks. A key component in its hardware deployment is the ability to generate dynamic reservoir states. Our research introduces a novel dual-memory RC system, integrating a short-term memory via a $WO_x$-based memristor, capable of achieving 16 distinct states encoded over 4 bits, and a long-term memory component using a $TiO_x$-based memristor within the readout layer. We thoroughly examine both memristor types and leverage the RC system to process temporal data sets. The performance of the proposed RC system is validated through two benchmark tasks: isolated spoken digit recognition and with only a fraction of complete samples forecasting the Mackey-Glass (MG) time series prediction. The system delivered an impressive 98.84% accuracy in speech digit recognition and sustained a low normalized root mean square error (NRMSE) of 0.036 in the time series prediction task, underscoring its capability. This study illuminates the adeptness of memristor-based RC systems in managing intricate temporal challenges, laying the groundwork for further innovations in neuromorphic computing.

## 1. Introduction

Synaptic connections within the human brain are instrumental in the intricate process of transmitting biological information, marking the foundation of neural communication. These connections enable the brain to differentiate between short-term memory (STM) and long-term memory (LTM) by dynamically modulating the synaptic weights, which in turn, regulates the strength of synaptic connections. This capacity for modulation underpins the brain's remarkable ability to adapt and learn. In the realm of computational architectures, the traditional von Neumann model shows limitations in handling extensive data volumes efficiently. This recognition has propelled research into artificial synapses, seeking to emulate the brain's efficiency and flexibility [1]. Neuromorphic computing systems, leveraging the principles of artificial neural networks (ANN), stand at the forefront of this exploration. These systems are uniquely poised to manage massive parallel processing tasks with minimal energy consumption, reflecting significant advancements in the field. They have catalyzed innovation across a spectrum of neurohybrid applications, ranging from sophisticated hardware prototypes for data classification to pattern recognition and beyond, into the realm of unsupervised learning [2–6]. The evolution of semiconductor memory devices has further expanded the possibilities for neuromorphic applications, offering new platforms for the realization of these advanced computational models [7]. Among the array of memory technologies explored, ferroelectric random-access memory (FeRAM) [8–9], spin-transfer torque magneto random access memory (STT-MRAM) [10–11], phase-change random access memory (PcRAM) [12–14], and resistive random-access memory (RRAM) [15–18] stand out. Each of these devices brings unique properties to the table, yet RRAM distinguishes itself through its simplicity, low energy requisites, fast switching capabilities, and seamless integration with conventional CMOS processing techniques [19–21]. This combination of attributes renders RRAM an optimal choice for the implementation of reservoir computing (RC) [22–25], a computational framework tailored for the nuanced processing of temporal and sequential data.

RC embodies a transformative approach in the realm of neural network training, characterized by its singular focus on the read-out layer. This architectural simplification, often harnessing techniques such as linear regression and classification, indicates RC's commitment to rapid learning and computational efficiency. The cornerstone of RC lies in its unique structure: a "reservoir" that augments input data into a higher-dimensional expanse, coupled with a read-out layer adept at interpreting intricate patterns within this transformed domain. Such a configuration circumvents the exhaustive training typically associated with conventional neural networks, confining the training process to the read-out layer alone, thereby promising a reduction in computational overhead [26]. The efficacy of RC hinges on two critical attributes: non-linearity and STM. Non-linearity is vital for transforming linearly inseparable inputs into states that the read-out layer can linearly unravel [27]. STM plays an equally crucial role, ensuring that the reservoir's output is predominantly influenced by recent inputs, which bolsters the system's responsiveness and temporal fidelity [28].

RC systems have demonstrated remarkable proficiency in tasks requiring an acute temporal data awareness. Their prowess in analyzing and forecasting time-sensitive patterns has been showcased across various disciplines, including speech recognition, financial forecasting, and environmental monitoring, underscoring their adaptability and broad applicability [29]. In the context of temporal data processing, recurrent neural networks (RNN) stand out with their neuron interconnectivity, which facilitates the integration of both present and historical input into their output. This feature allows RNN to extract temporal patterns and dependencies that are beyond the reach of traditional feedforward networks [30]. The integration of RRAM into RC systems has been a pivotal step in neuromorphic computing. RRAM devices fulfill the critical requirements for effective physical implementation of RC: non-linearity for processing complex inputs and STM for enhancing system reactivity [27]. Moreover, the RRAM's ability to mimic the synaptic functions within neural circuits complements the RC framework, which inherently seeks to replicate the brain's operational principles. The synergy between RC and RRAM could lead to the creation of more efficient, scalable, and adaptable computing paradigms. By emulating the brain's efficiency and flexibility, incorporating RRAM into RC systems represents a significant leap towards computing technologies that align more closely with biological neural processes [31-34]. This integration underscores a concerted effort in computational neuroscience to harness the advantages of physical systems for advanced computing architectures. It encapsulates the ambition to craft artificial synapses, advancing the frontier of neuromorphic computing systems that combine the dynamism of RC with the innate properties of memory semiconductors.

In our study, we introduce a cutting-edge memristor-based RC framework tailored for data-intensive computing tasks, such as neural network training, which are essential for artificial intelligence applications but often involve significant energy consumption. Our approach utilizes specialized hardware to directly map neural networks, leveraging arrays of memristive devices for parallel multiply–accumulate operations. We present a novel memristive device, validated using the Cadence tool, and developed a Python-based model to facilitate the design of a memristive -based accelerator. This framework employs a crossbar array of memristive devices to train a neural network for digit classification and time series prediction. Our system effectively addresses the non-ideal characteristics of memristive devices, achieving remarkable training accuracies of 98.84% for spoken-digit recognition but also maintained a low normalized root mean square error (NRMSE) of 0.036 in time series prediction, showcasing its proficiency. This research demonstrates the efficacy of memristive-based neural networks in handling classification tasks, showcasing their potential in reducing the energy demands of data-intensive computing. The integration of memristive technology within neural networks not only enhances computational efficiency but also underscores the viability of these systems in neuromorphic computing. Our work sets the stage for further advancements by illustrating the practical applications of memristive in neural network training and classification tasks. The significant training accuracies achieved highlight the promise of memristive -based systems in AI, suggesting a substantial impact on future developments in energy-efficient and high-performance computing.

## 2. Memristor Model and Reservoir Framework

### 2.1 $TiO_x$ for NVM Memristor and $WO_x$ for DM

We delve into the realm of memristive technologies, spotlighting NVM memristors, particularly those utilizing $TiO_x$ as their core material. These devices, along with other variants like AIST-based [35] and PCMO-based memristors [36], are at the cutting edge of advancements in memristive technology, offering novel solutions in the fabrication of neuromorphic hardware for ANN. The dynamic adjustment of memristor conductance in response to voltage variations draws a parallel to the behaviour of biological synapses, which adapt based on neuronal activity, positioning the memristor as a pivotal element in neuromorphic computing architectures. To capture the dynamic properties of memristors within simulation environments accurately, a plethora of SPICE models have been introduced, including the TEAM [37] and VTEAM models [38], alongside specific ones tailored for AIST, $WO_x$, and $TiO_x$ memristors. This work employs two distinct types of memristors: DM for creating a physical reservoir in the temporal data reservoir (TDR) [30] and NVM paired with resistors for energy-efficient computational tasks.

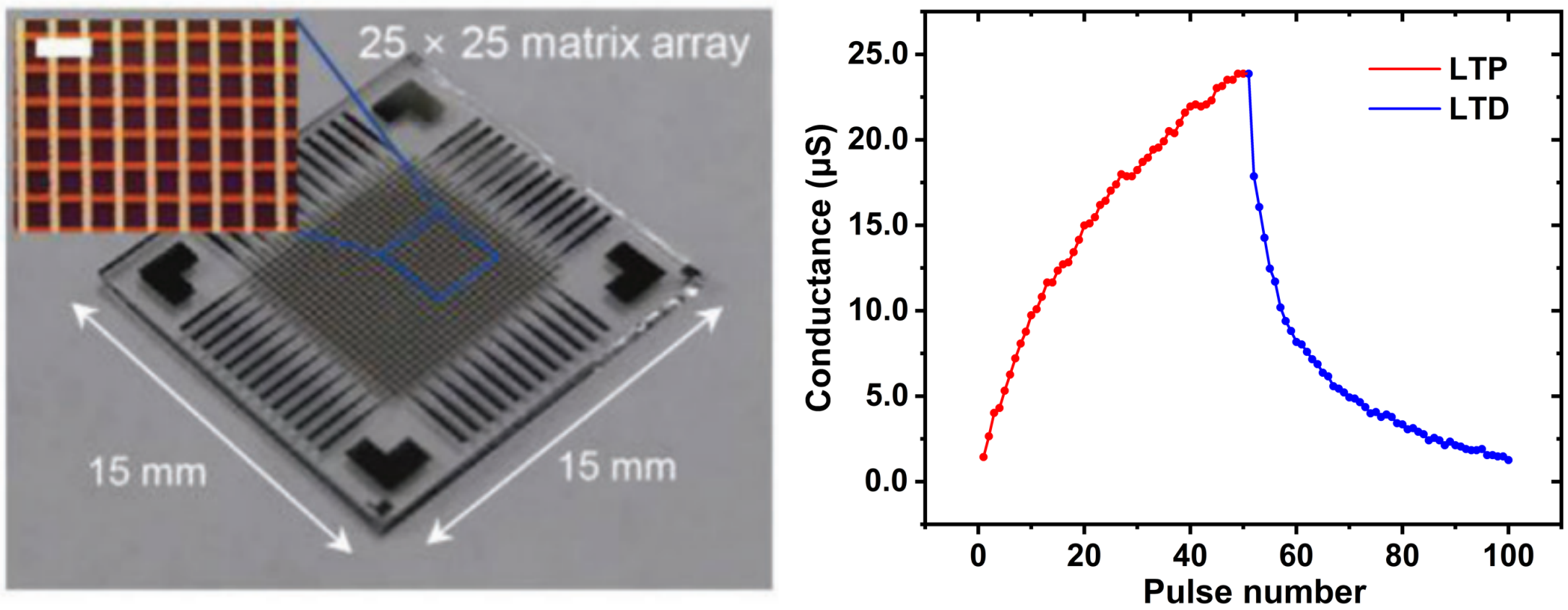

Fig. 1. Experimental (a) $TiO_x$ memristor array and (b) Conductance change of the $TiO_x$ device.

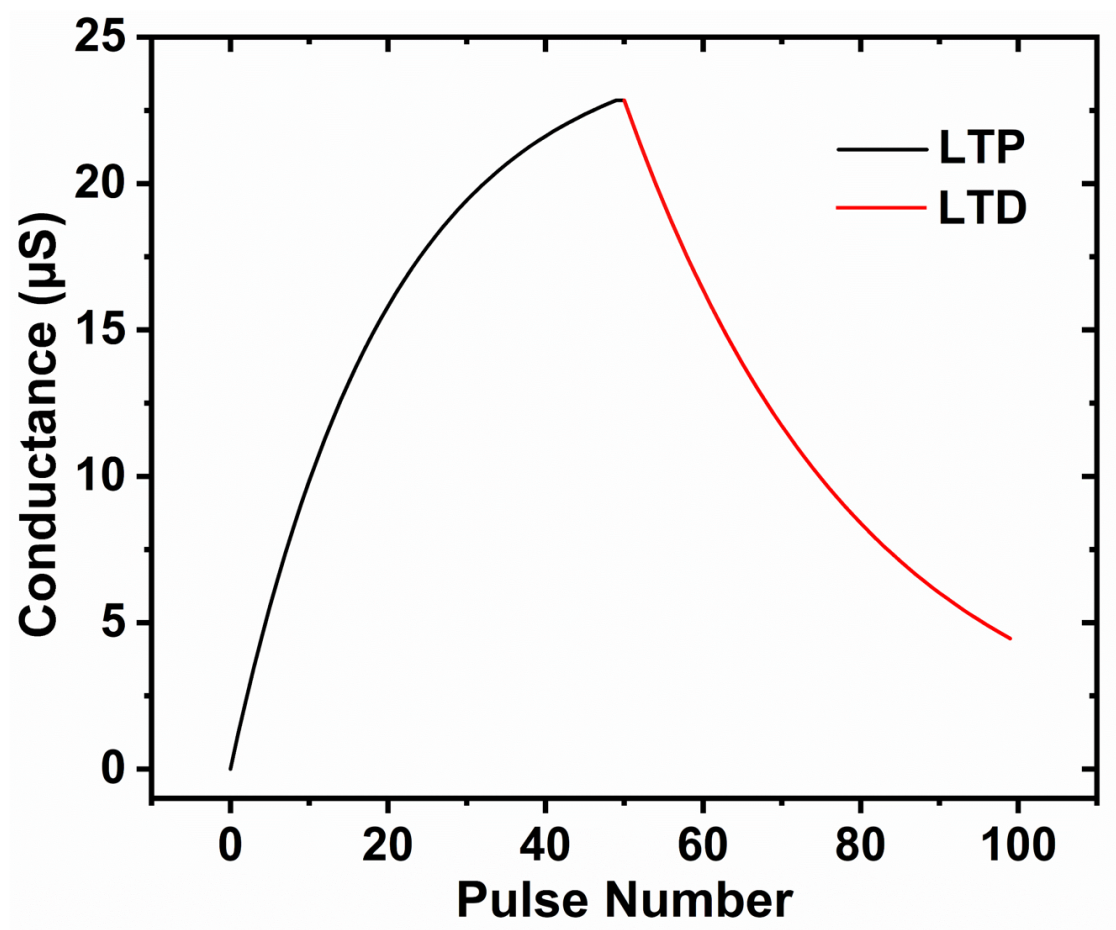

Fig. 2. Proposed framework conductance change for the $TiO_x$ device.

The choice of $WO_x$ and $TiO_x$ models to simulate the dynamic characteristics of DM and NVM highlights our commitment to accurately depict the complex behaviours of memristive devices. Our analysis indicates that traditional models, such as the HP model, are insufficient for describing the nuanced dynamics of $TiO_x$-based NVM memristors and their equivalents. This inadequacy has led to the development of a new model that aligns with the observed behaviours of contemporary memristive devices, marking a substantial advancement in memristor simulation. This novel model significantly enhances our understanding and application of $TiO_x$-based NVM memristors, paving the way for creating more advanced and efficient neuromorphic systems. Through this work, the $WO_x$ and $TiO_x$ models are specifically selected to simulate the dynamic characteristics of DM and NVM, respectively, addressing the derivative of the state variable *w(t)* within the $TiO_x$ memristor model [39].

$$\frac{dw(t)}{dt} = \begin{cases} \mu_v \frac{R_{ON}}{D} \frac{i_{OFF}}{i(t) - i_O} f(w(t)), & v(t) > V_{T+} > 0 \\ 0, & V_{T-} \le v(t) \le V_{T+} \\ \mu_v \frac{R_{ON}}{D} \frac{i(t)}{i_{ON}} f(w(t)), & v(t) < V_{T-} < 0 \end{cases} \tag{1}$$

$$f(w(t)) = 1 - \left( \frac{2w(t)}{D} - 1 \right)^{2p} \tag{2}$$

The symbol $\mu_v$ represents the average mobility of ions, a fundamental property influencing the device's operational dynamics. The term $R_{ON}$ signifies the memristor's minimal resistance or memristance, establishing a baseline for its performance capabilities. The constants $i_{OFF}$, $i_{ON}$, $p$, and $i_O$ are critical in fine-tuning the model to align with observed behaviours and theoretical frameworks. These parameters are instrumental in precisely representing the memristor's electrical response under varying conditions. The memristor's physical dimension, particularly its thickness denoted by $D$, directly impacts its electrical properties and is thus a vital factor in the model. Threshold voltages, $V_{T+}$ and $V_{T-}$, set the operational boundaries for the device, defining the voltage range within which the memristor functions optimally and safely. Including a window function, $f(\cdot)$, is a safeguard, ensuring that the state variable remains within a specific range, thus preventing operational extremes that could compromise the device's integrity.

Figure 1(a) presents an optical microscopic and a photographic image of a 25 x 25 passive crossbar array, which incorporates $TiO_x$-based memristor nodes. Situated at the intersections of vertically arranged Al electrode lines, each with a width of 100 µm, the memristor cells embody an Al/TiOx/Al layered structure atop a 15 mm x 15 mm glass substrate. This configuration is pivotal for its application as a matrix component in artificial neural networks, where it functions as a two-terminal artificial synapse that bridges pre- and post-synaptic neurons. The post-synaptic current in each memristor cell is delicately adjustable, responding to the history of electrical pulse spikes emanating from the presynaptic neuron. This feature is critical to the crossbar array's ability to facilitate learning and memory processes within a neural network structure. Figure 1(b) illustrates the conductance response of the memristor device subjected to 50 SET pulses (2.5 V, 100 ms) and 50 RESET pulses (−2.5 V, 100 ms). The response showcases the device's intrinsic asymmetry in conductance change, distinguishing between long-term potentiation (LTP) and long-term depression (LTD), which are fundamental for synaptic plasticity emulation.

In our framework, we have emulated the behaviour of a single $TiO_x$ memristor using the parameters shown in Table 1 for the implementation of the readout layer. This adaptation allows the memristor's conductance to be dynamically adjusted within the framework, catering to the application's specific requirements and the dataset's complexity, thereby enhancing accuracy. Figure 2 showcases this adaptation, illustrating the relationship between the pulses applied to the device and the resulting changes in conductance, closely aligning with the conductance behaviour observed in the actual fabricated data. Furthermore, we have extended the application of the same memristor model to investigate its response under a sinusoidal voltage stimulus. By monitoring the current in response to this stimulus, we have plotted a hysteresis curve that elucidates the relationship between voltage and current and shown in Figure 3(a) and (b). This exploration not only underscores the versatility of the $TiO_x$ memristor in simulating complex neural network operations but also provides valuable insights into its electrical characteristics under varying input conditions.

TABLE 1. Parameters of $TiO_x$ NVM model.

| **Parameters** | ***$V_{T+}$ (V)*** | ***$V_{T-}$ (V)*** | ***D (nm)*** | ***$\mu_v$*** | ***$I_{on}$ (A)*** | ***$I_{off}$ (A)*** | ***io(A)*** | ***p*** |
|---|---|---|---|---|---|---|---|---|
| **Values** | 1 | -1 | 10 | 10e-17 | 20e-6 | 22e-6 | 1e-6 | 10 |

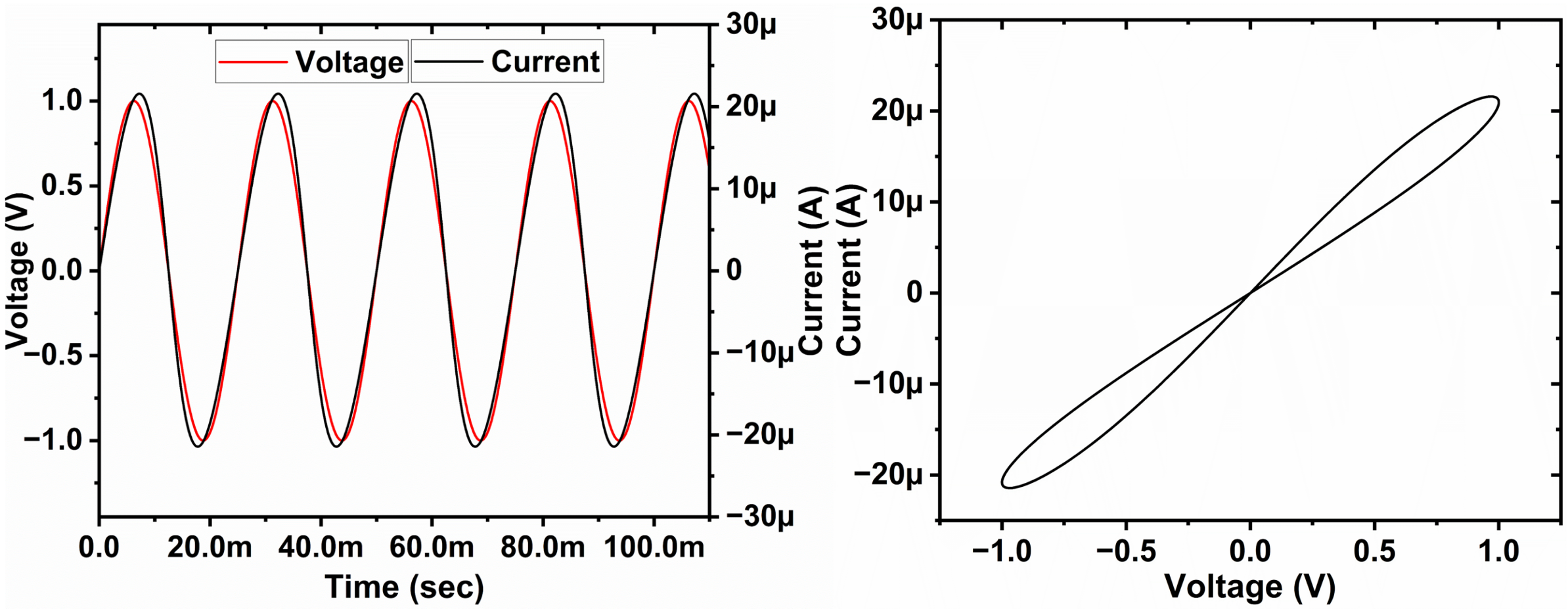

Fig. 3. Proposed framework simulation with $TiO_x$ memristor (a) Transient curve and (b) hysteresis curve.

Moreover, the $WO_x$ memristor model, [40], can be elucidated as follows:

$$I(t) = (1 - w(t))\alpha\left[1 - e^{-\beta . V(t)}\right] + w(t)\gamma \sinh(\delta V(t))$$
$$\frac{dw(t)}{dt} = \lambda \sinh(\eta V(t)) - \frac{w(t)}{\tau} \quad (3)$$

In the given equation, *I(t)*, *V(t)*, and *w(t)* represent the current, voltage, and state variables of the memristor, respectively. The parameters $\alpha$, $\beta$, $\gamma$, $\delta$, $\lambda$, and $\eta$ are all positive and dependent on the material properties of the memristor. The symbol $\tau$ represents the diffusion time constant, a critical parameter influencing the performance of the designed RC system. Eq. (3) clearly demonstrates a pronounced nonlinear relationship between the current and voltage in the $WO_x$ memristor. This nonlinearity is crucial for the reservoir's functionality, as it allows for the original input signal to be mapped into a high-dimensional state space, thereby facilitating complex computations.

In our framework, we utilized data from a fabricated $WO_x$ memristive crossbar array, which consists of memristors structured in a Pt/$WO_x$/W stack. Figure 4(a) displays a top-view image of the 16 x 16 array. The precise engineering of this array is crucial for its deployment in advanced neuromorphic computing systems, where the exact dimensions of the memristive components are essential for precise conductance modulation. We have emulated the behaviour of a single $WO_x$ memristor using the parameters shown in Table 2 to implement the reservoir layer. Figure 4(b) illustrates the experimental evaluation of the memristor's ability to discriminate between sixteen unique pulse stream inputs. Each input stream alters the memristor's state, which is quantitatively evidenced by the corresponding read current measurements taken after pulse application. This ability to differentiate input patterns is critical for complex signal-processing applications. The memristor's response to varied pulse streams was systematically examined, as depicted in Figure 4(b). Upon application of each unique pulse stream, a distinct memristor state emerged, as indicated by the read current. This behaviour confirms the memristor's capability to differentiate among sixteen discrete pulse stream patterns. For the encoding of a binary '1', a set pulse of 1.8 V was utilized, and the conductance states were then read with a 0.5 V pulse to ensure a 4-bit resolution. The memristor's conductance range, spanning approximately 0.5 mS to 0.8 mS, allows for the clear distinction between binary '1' and '0', enabling the creation of all 16 unique states. The ability of the memristor to modulate conductance and differentiate states, as presented in Figure 4(b), underscores its applicability in multi-state logic operations within neuromorphic computing architectures. Figure 5 showcases this adaptation, illustrating the relationship between the pulse streams to the device and the resulting changes in conductance, closely aligning with the conductance behaviour observed in the actual fabricated data.

TABLE 2. Parameters of $WO_x$ DM model.

| **Parameters** | $\alpha$ | $\beta$ | $\gamma$ | $\delta$ | $\lambda$ | $\eta$ | $\tau$ |
|---|---|---|---|---|---|---|---|
| **Values** | 2.5e-6 | 0.5 | 2.5e-6 | 4 | 2.5 | 2 | 0.05 |

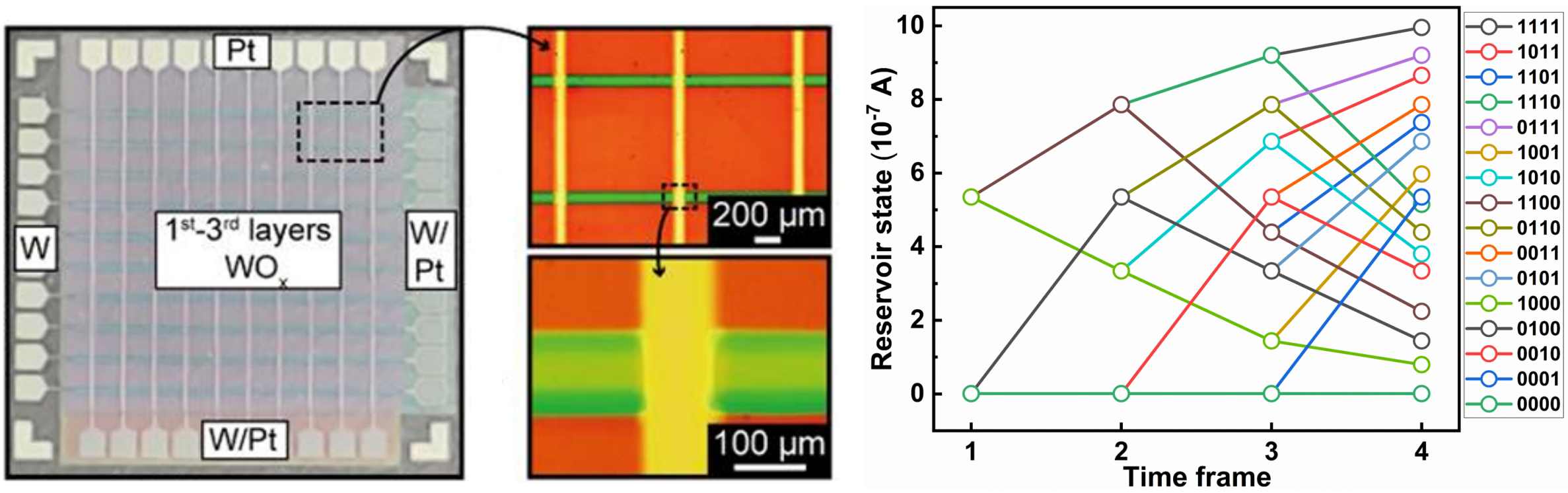

Fig. 4. Experimental (a) $WO_x$ memristor array and (b) Conductance change of the $WO_x$ device.

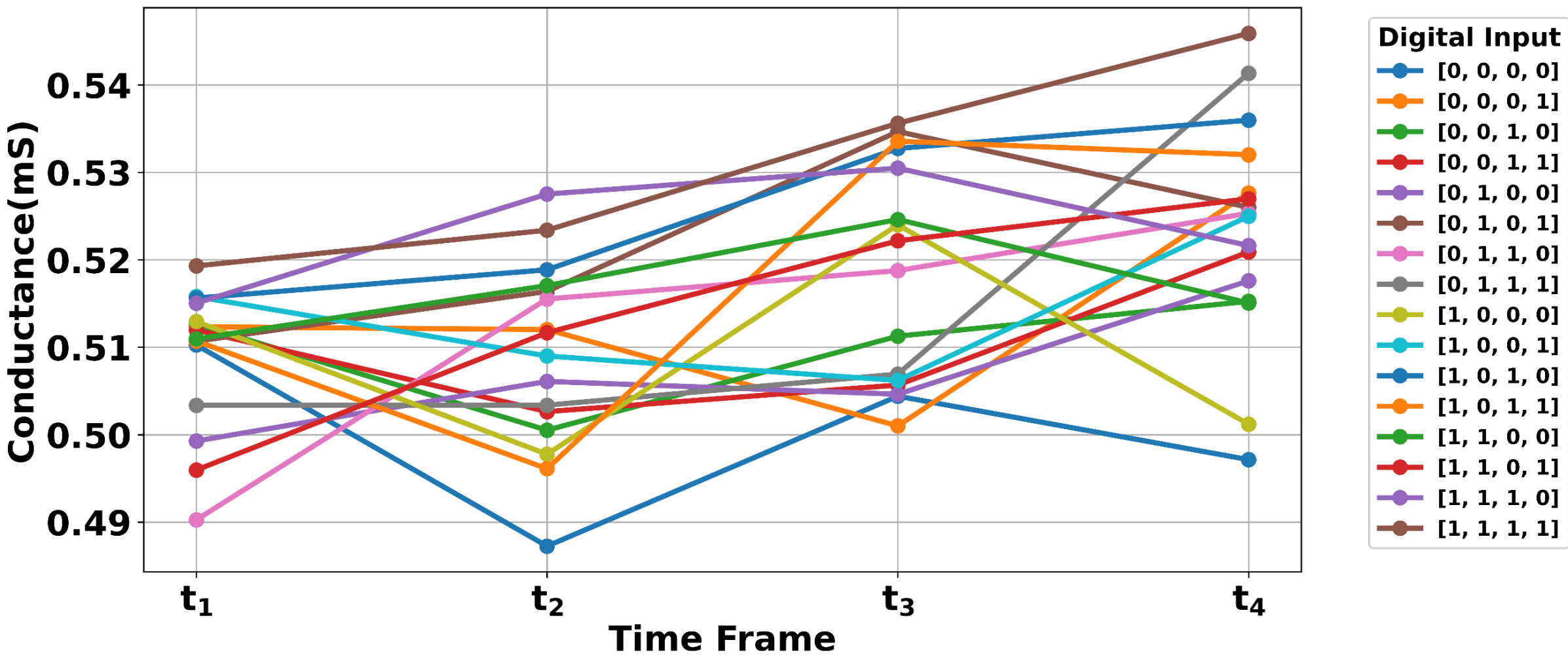

Fig. 5. Proposed framework conductance change of the $WO_x$ device.

### 2.2. Reservoir Framework

The conventional RC architecture typically comprises three distinct layers: an input layer, a reservoir, and an output layer, as depicted in Fig. 6. The reservoir layer is characterised by a dense network of sparsely connected neurons. These neurons can nonlinearly transform the input signal into a high-dimensional state space, enhancing the reservoir states' linear separability. The equation governing the update of neuronal states within the reservoir layer is as follows:

$$x(n+1) = f(W^{in}u(n) + W^{res}x(n) + W^{back}\hat{y}(n) + b) \tag{4}$$

In this formulation, $x(n+1)$ represents the state vector of the reservoir at time step $n+1$. The function $f(\cdot)$ denotes a nonlinear activation function that is applied element-wise, examples of which include the hyperbolic tangent function (*tanh*). The matrix $W_{in}$ constitutes the weight matrix governing the connections from the input to the reservoir, while $u(n)$ is the vector representing the input at time step $n$. Furthermore, $W_{res}$ is the weight matrix responsible for the internal connections within the reservoir itself. The state vector of the reservoir at time step $n$ is denoted by $x(n)$. Additionally, $W_{back}$ represents the weight matrix for the feedback connections, which convey the output back to the reservoir. The predicted output vector at time step $n$ is represented by $\hat{y}(n)$, and $b$ signifies the bias vector. This structure is integral to the operation of echo state networks (ESN), a class of RNN, where the dynamical properties of the reservoir facilitate the processing of temporal sequences through a rich, high-dimensional representation of the input data. The reservoir's dynamics, modulated by the specified weight matrices and the nonlinear activation function, enable the network to capture and exploit temporal correlations in the input data for tasks such as prediction, classification, and generation of temporal sequences.

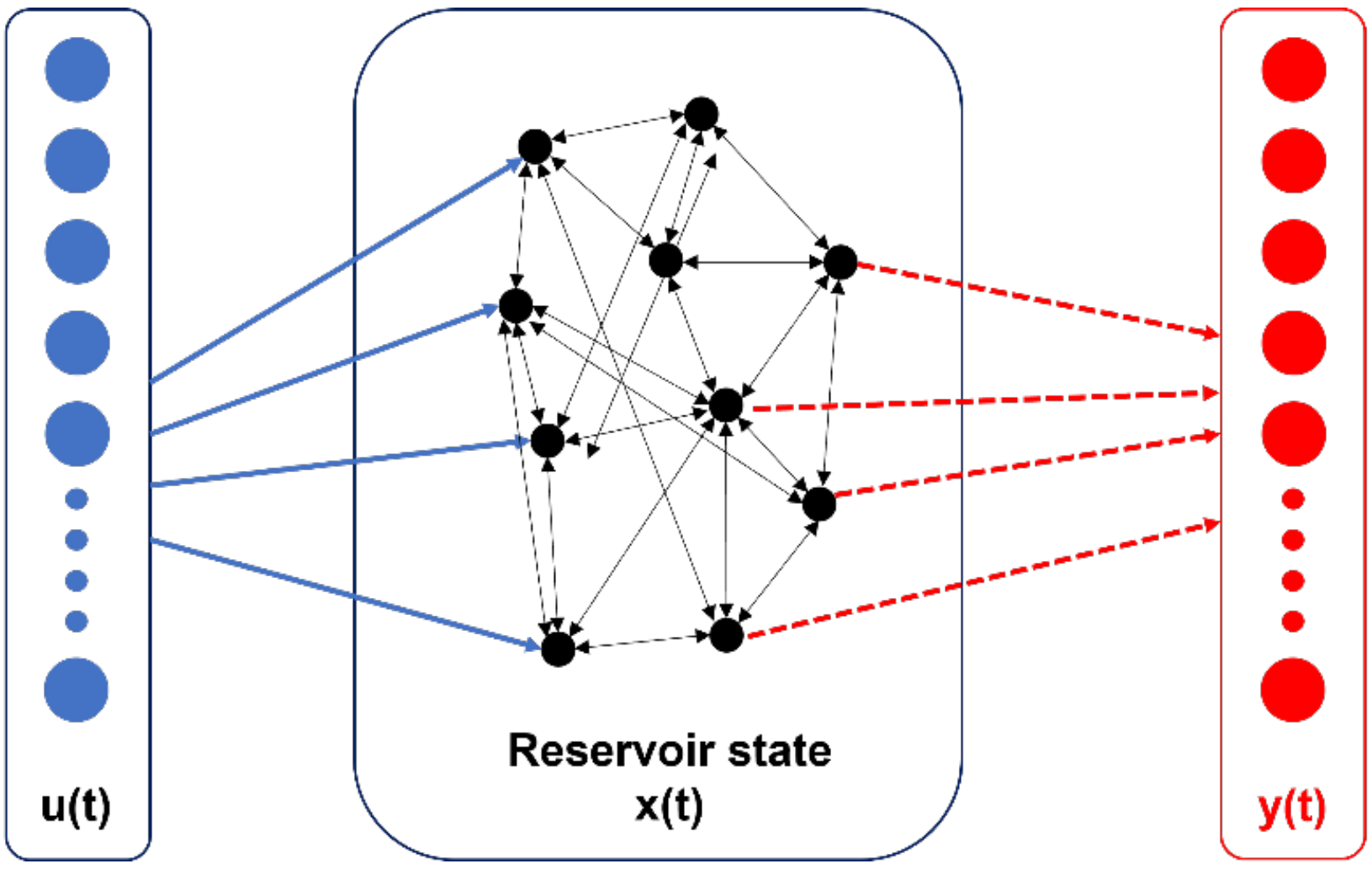

Fig. 6. Schematic of a conventional RC system.

Implementing large-scale analog reservoirs in hardware presents significant challenges due to the requirement for a substantial number of neuron nodes within the reservoir. Additionally, the variability in the characteristics of non-volatile memristors, which are employed as weight computing units, complicates circuit implementation when dealing with extensive weight parameters. To address these issues, the concept of a TDR utilizing a single dynamic node has been proposed. This innovative approach simplifies the hardware architecture while mitigating the impact of device-to-device variations [41]. The schematic of the TDR system is depicted in Fig. 6, and its operational model is delineated as follows:

$$\hat{x}(t) = F(x(t), x(t-\tau) + \gamma I(t) M(t)) \tag{5}$$

In the TDR model, *F* represents the underlying dynamical system, characterized by a feedback loop delay denoted as $\tau$, and a scaling factor $\gamma$. The function *I(t)* signifies the original input signal, while *M(t)* describes a mask function. This mask function is defined as a piecewise constant function with a period of $\tau$ and maintains consistency over an interval of $\theta$ [42]. The values of the mask function are randomly generated from a specified probability distribution, ensuring a diverse temporal modulation of the input signal. This configuration diverges from traditional RC architectures, which typically rely on creating a vast network of real nodes. Instead, the TDR approach requires only a single physical node, such as a memristor device, to implement the entire reservoir circuit. This novel strategy substantially mitigates the impact of device-to-device variations inherent in memristor technologies, thereby enhancing the stability and reliability of the reservoir's performance. The TDR model offers a promising avenue for implementing efficient and scalable RC systems by reducing the complexity and susceptibility to variations.

## 3. Memristive-based Fully Reservoir System

In this section, we introduce a framework for a fully memristive RC circuit, establishing a new standard in the efficient real-time processing of spatiotemporal information. This development represents a substantial leap forward from the capabilities of conventional RNN. Central to our work is a meticulously detailed schematic diagram illustrated in Fig. 7, showcasing the deployment of multiple $WO_x$-based digital memories in a cascading configuration [30]. This design forms the backbone of our reservoir circuit, enabling a hierarchical approach to processing spatiotemporal signals. To complement this, we have integrated a series of parallel reservoir circuits, each equipped with its own distinct masking process, thus optimizing the overall architecture's effectiveness. The innovation extends to the readout layer, which utilizes $TiO_x$-based memristor technology for vector-matrix multiplication (VMM), a critical step in translating the processed signals into meaningful outputs.

The architecture of the circuit is thoughtfully divided into three principal components: the input module, the reservoir module, and the readout module. The initiation of the process occurs at the input module, which is tasked with generating the initial input voltage in conjunction with mask pulse voltages. These elements are derived from datasets and mask sequences that have been implemented in python with the provided algorithm, demonstrating the framework's computational efficiency and adaptability. This framework highlights the potential of memristive components in revolutionizing RC circuit design and opens new avenues for research into real-time processing of complex data structures, making it a valuable contribution to the field of neuromorphic computing.

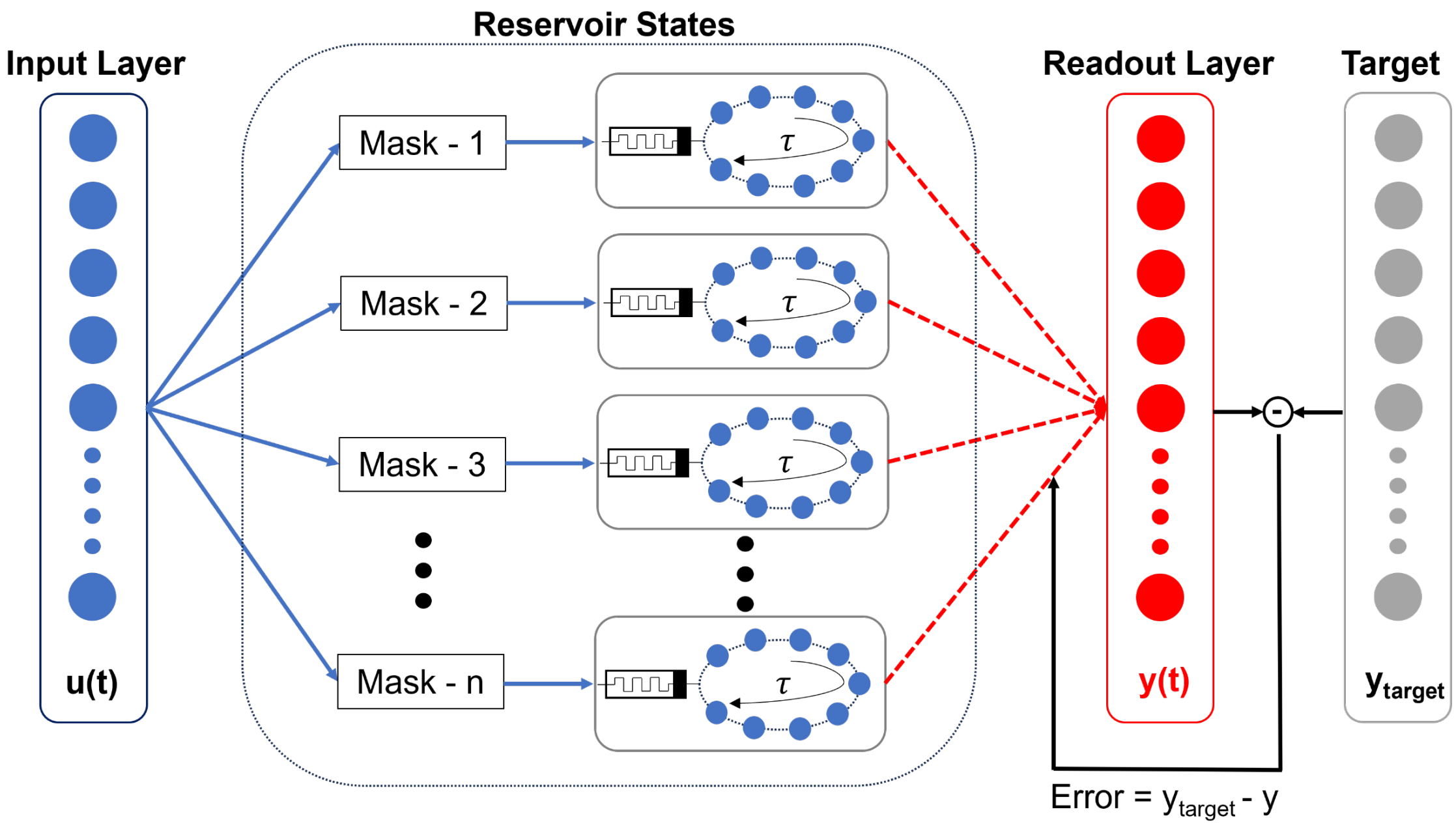

Fig. 7. Schematic of the dynamic memristor-based RC system.

### 3.1 Input Layer

The input module of the described RC framework, as shown in Figure 8, is adeptly architected to handle a broad range of temporal signals, not limited to audio data or any specific dataset. This generalization allows the framework to be applicable across various domains where temporal dynamics are key, such as speech recognition, environmental sound classification, biomedical signal analysis, and more. We have used MG time series forecasting and speech recognition in our framework.

Initially, the process commences with the loading and resampling of temporal signals to a uniform sampling rate, establishing a consistent basis for all subsequent analyses. This uniformity is essential for eliminating potential discrepancies arising from varied sampling rates across the dataset. Following this, the module shifts focus to the extraction of domain-specific features. This step is adaptable, with the extraction technique tailored to the nature of the signal; for instance, audio signals may be processed using Mel-Frequency Cepstral Coefficients (MFCC) to capture essential characteristics like timbre and pitch, whereas other signals might necessitate different features such as power spectral density or wavelet transforms. The essence of this phase is to distil the most informative attributes of the signal pertinent to the task at hand. Post-feature extraction, the input data undergoes normalization to fit within a designated range, a prerequisite for ensuring stable dynamics within the reservoir. Subsequently, a dimensionality expansion is performed, often through random masking or projection, to enhance the diversity and separability of input patterns. This expansion is a critical facilitator for the dynamic processing elements within the reservoir, such as memristors, enabling them to effectively capture and leverage complex temporal dependencies inherent in the input data. Finally, an optional yet beneficial step involves explicitly encoding temporal dynamics into the processed signal. Techniques like delay embedding or differential encoding may be employed to enrich the input representation with temporal information, thereby providing a more comprehensive input to the reservoir. This enriched input enables the reservoir to understand better and process the temporal evolution of signals, significantly augmenting its computational prowess across diverse applications. Through these meticulously designed steps, the input module ensures that the RC system is adeptly equipped to handle a vast array of temporal signals, effectively translating intricate temporal dynamics into a structured format conducive to advanced computational analysis.

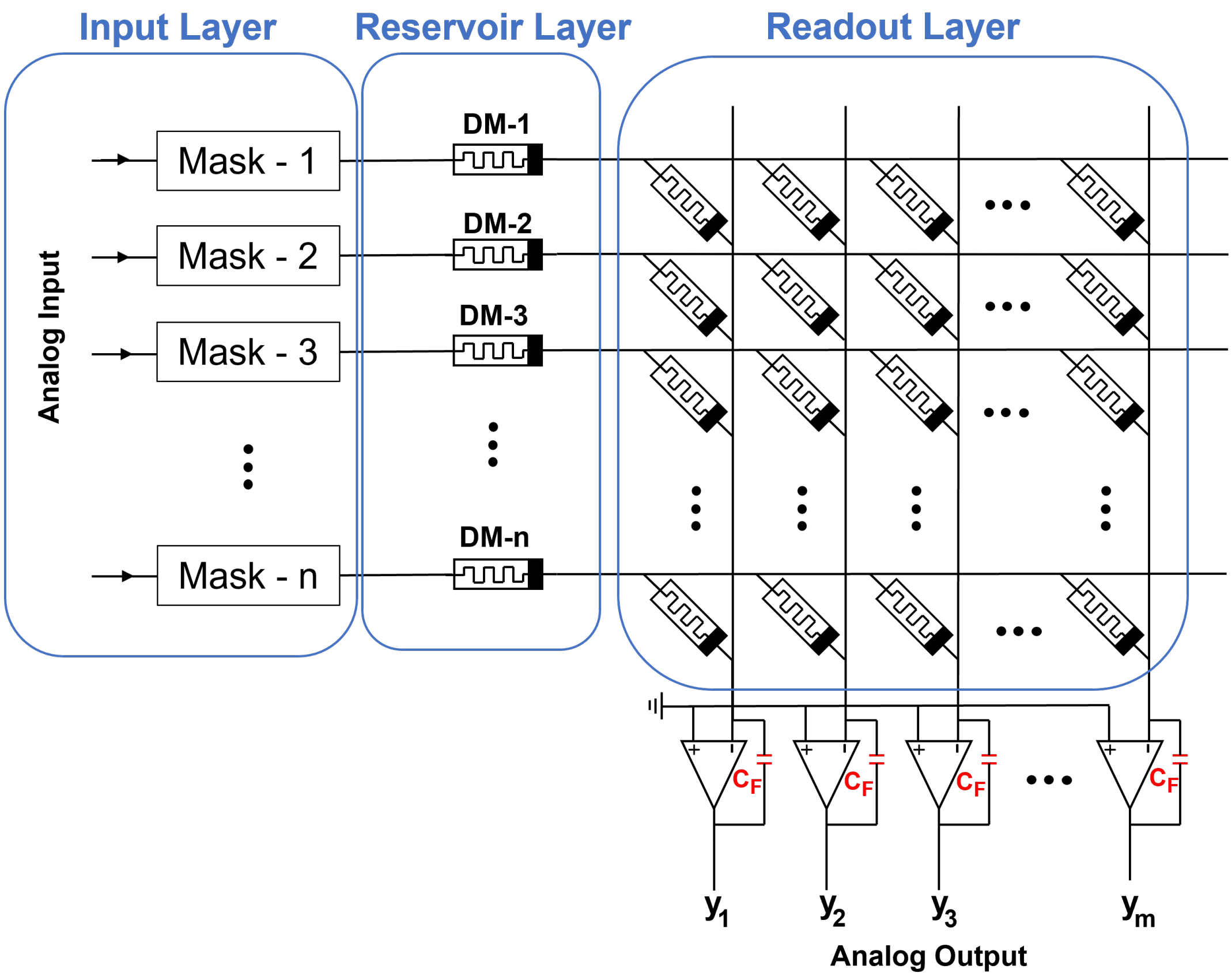

Fig. 8. The proposed schematic outlines an RC circuit that features four distinct modules: an input module, a mask module, a reservoir module that utilizes $WO_x$ memristors, and a readout module designed with a $TiO_x$-based memristor crossbar network.

### 3.2 Reservoir Layer

In the DM-based RC framework presented in Fig. 8, the operational essence is captured by the sophisticated implementation of reservoir states, which is crucial for the adept decoding of temporal data. This implementation hinges on leveraging the intrinsic properties of DM, which exhibit variable conductance in response to applied voltage stimuli, thereby encoding a memory of past inputs. The conductance change of each memristor, governed by nonlinear dynamics, facilitates a unique mapping of input signals into a high-dimensional state space. This mapping is achieved through a preliminary mask layer that disperses input features across the memristor array, ensuring a diversified response pattern that enhances the system's sensitivity to varying data patterns. The reservoir's transformation process is pivotal; upon interacting with the mask layer and subsequent memristor array, each input vector undergoes a transformation into a reservoir state vector. This vector is a high-dimensional representation imbued with temporal context, courtesy of the memristors' dynamic conductance states. These states effectively capture and encode temporal dependencies inherent in the input data, laying a robust foundation for the subsequent readout layer. This readout layer, the sole trained component of the system, combines the enriched reservoir states to produce the final output, showcasing the system's efficiency and the minimalistic nature of reservoir computing.

The adoption of DM for encoding reservoir states underscores the system's efficiency and adaptability and mirrors the operational dynamics of biological neural networks. This resemblance is attributed to the memristors' nonlinearity and capacity for adaptive learning, enabling the processing of complex temporal patterns with remarkable efficiency. Consequently, the system's design, centred around DM-based reservoir states, marks a significant advancement in computational models, particularly for applications necessitating nuanced temporal data processing and decoding.

**Algorithm:** Analog memristive reservoir computing for temporal data classification

```
Input: Dataset of temporal data, DM and NVM parameters, number of DM (N) and NVM, mask length
Output: Trained model, Accuracy, NRMSE, weights of the NVM VMM array

1: Initialize:
  - Load the dataset
  - Extract features
  - Encode labels with One-Hot Encoding
  - Split dataset into training and testing sets

2: Create a random bipolar mask with dimensions (N, ML)

3: Standardize the feature set using StandardScaler

4: Normalize input with a specified voltage range (Vmax, Vmin)

5: Initialize memristor conductance array G

6: Define DM function

7. Define NVM function

8: for each sample in the training set do
    - Apply the mask to the sample and obtain the input voltage
    - Calculate the memristor output current using the dynamic memristor model
    - Store the memristor current in reservoir states
  end for

9: Repeat step 7 for the test set to obtain reservoir states

10: Define the neural network model with layers:
  - MemristorCrossbarLayer for dimensionality expansion
  - Activation layer with ReLU function

11: Compile the model with Adam optimizer and categorical crossentropy loss

12: Train the model on the training set reservoir states and labels

13: Evaluate the model on the test set reservoir states and labels

14: Predict labels for the test set, NRMSE and calculate test accuracy

15: Output the trained model, NRMSE and test accuracy
```

### 3.3 Readout Layer

In the proposed RC framework, the readout layer is pivotal in deciphering temporal datasets through a VMM strategy shown in Fig. 8, underpinned by a sophisticated $TiO_x$ memristor model. This model provides a comprehensive understanding of memristive dynamics, offering an advanced means for processing and interpreting temporal information. The essential function of the readout layer is to linearly amalgamate the reservoir's high-dimensional states, culminating in an output that precisely reflects the temporal patterns observed in the input data. The $TiO_x$ memristor model is distinguished by its unique memristive characteristics, including state-dependent conductance and the capacity to mimic synaptic plasticity, laying a solid foundation for the readout mechanism. It encompasses key parameters such as the memristor's physical dimensions, resistance states, and charge mobility, crucial for articulating its dynamic response to electrical stimuli. Utilizing these attributes, the $TiO_x$ memristor enables a VMM operation wherein the input vector—representing the reservoir states—is processed through a memristive crossbar array, thus facilitating the readout function. The VMM can be represented mathematically as:

$$\left[ y_1 \cdots y_m \right] = \left[ DM_1 \cdots DM_n \right] \bullet \begin{bmatrix} w_{11} & \cdots & w_{1m} \\ \vdots & \ddots & \vdots \\ w_{n1} & \cdots & w_{nm} \end{bmatrix} \tag{6}$$

The equation delineates the VMM process within an RC framework utilizing DM. The resulting vector [$y_1 \ldots y_m$] comprises '$m$' output elements of the VMM operation. The matrix on the right, represented as [$DM_1 \ldots DM_n$], consists of '$n$' DM columns corresponding to the outputs of the reservoir layer. This matrix configuration is defined by 'm' rows and '$n$' columns, where each entry $w_{ij}$ signifies the synaptic weight connecting the '$i$-th' output to the '$j$-th' input node. Through the VMM operation, an '$n$'-dimensional input vector is transformed into an '$m$'-dimensional output vector, where each output component $y_i$ is computed as the sum of the products of the corresponding synaptic weights $w_{ij}$ and the input nodes $DM_j$, for '$j$' ranging from $1$ to '$n$'.

Implementing the readout layer via the $TiO_x$ memristor model entails translating the reservoir's output states across a memristive crossbar. Here, the conductance of each memristor embodies a synaptic weight. These weights, reflective of the $TiO_x$ memristor's intrinsic properties and their interaction with input signals, are fixed and emulate the dynamic adjustment of synaptic connection strengths seen in biological synapses, thereby fostering learning and adaptability within the system. The VMM task performed by the readout layer efficiently maps the temporal features encoded by the reservoir into a format suitable for subsequent classification or prediction tasks. This efficiency is achieved by multiplying the reservoir's high-dimensional output states against the conductance matrix of the memristive crossbar (representing synaptic weights), linearly integrating the states to yield the final output. The $TiO_x$ memristor model's adeptness at facilitating VMM operations significantly bolsters the system's proficiency in accurately decoding complex temporal patterns.

Incorporating the $TiO_x$ memristor within the readout layer presents numerous benefits, such as reduced power consumption, compact design, and the capability for parallel signal processing, mirroring the human brain's similar processing functions. Additionally, the dynamic modulation of memristive conductance states imparts a level of plasticity to the system, which is essential for learning and memory functions in neuromorphic computing.

## 4. Results and Discussion

### 4.1 Speech Recognition

The Free-Spoken Digit Dataset (FSDD) is a publicly available collection of high-quality voice recordings designed to advance research in speech recognition and related fields [48]. It includes WAV format recordings of spoken digits (0 to 9) in English, articulated by various speakers. This diversity enriches the dataset, making it an indispensable tool for developing and evaluating speech recognition models, particularly those aimed at identifying spoken numbers. The FSDD's recordings, typically sampled at 16 kHz, ensure the audio's fidelity, facilitating effective processing. The variety of speaker characteristics, such as pitch, accent, and speed, enhances the robustness of models trained on this dataset, positioning FSDD as an ideal resource for voice-based machine-learning explorations. The input module plays a pivotal role in processing FSDD's audio files. Utilizing the librosa library, this module loads each audio file and computes MFCC, which serves as a compact representation of the sound's short-term power spectrum. This critical step transforms the raw audio data into a structured form that is amenable to machine learning algorithms. The preprocessing phase further includes normalization and standardization techniques, standardizing the input data's scale and distribution. These preprocessing measures are vital for optimizing the learning efficiency of the model, thereby enhancing its predictive accuracy. Our proposed system utilizes a $WO_x$ DM-based reservoir module to process complex speech signal dynamics, generating high-dimensional reservoir states that serve as features for learning. These memristors, with their nonlinear dynamics and memory retention, are pivotal in capturing temporal patterns in speech. A $TiO_x$ memristor crossbar array forms the readout layer, transforming reservoir states for digit classification. Trainable weights within this layer adapt over multiple epochs, enhancing the model's predictive accuracy. This structured training enables the model to finely tune its recognition capabilities, demonstrating the potential of memristor-based neural networks for efficient speech recognition tasks.

We utilized an array of 40 DM to capture reservoir states, complemented by a 40 x 64 NVM memristor crossbar array that constitutes the readout layer. The training phase for the readout layer spanned 200 epochs, resulting in a high training accuracy of 98.84% based on a set of 2250 samples. The validation accuracy was also notable, reaching 93.87% with a subset of 750 samples. The size of the sample set is adjustable to meet the specific requirements of different applications. The progression of model accuracy over the epochs is graphically depicted in Figure 9(a). Additionally, Figure 9(b) provides a visual representation of the confusion matrix, illustrating the correlation between the true labels and the predicted labels, thereby offering insights into the model's performance. Following the comprehensive training of the readout layer, Figure 10 showcases the optimized weights within the memristor crossbar array. These weights are indicative of the learned patterns that enable the model to perform with high accuracy. This work not only demonstrates the effectiveness of memristor-based RC in dealing with complex temporal tasks but also paves the way for future innovations in analog computing architectures.

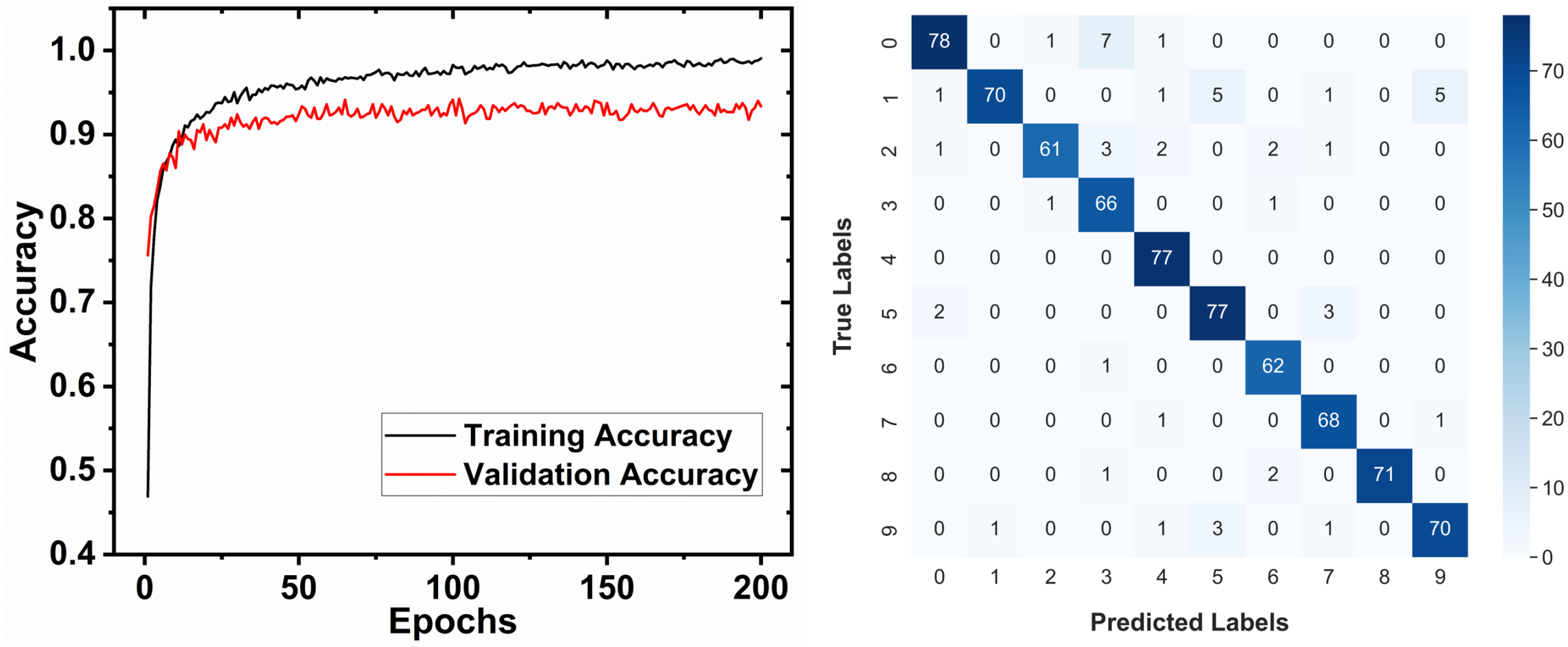

(a) (b)

Fig. 9. (a) Evaluating Model Robustness in Speech Recognition: Training and Validation Accuracy per Epoch (b) Confusion Matrix of Model Predictions Against True Labels.

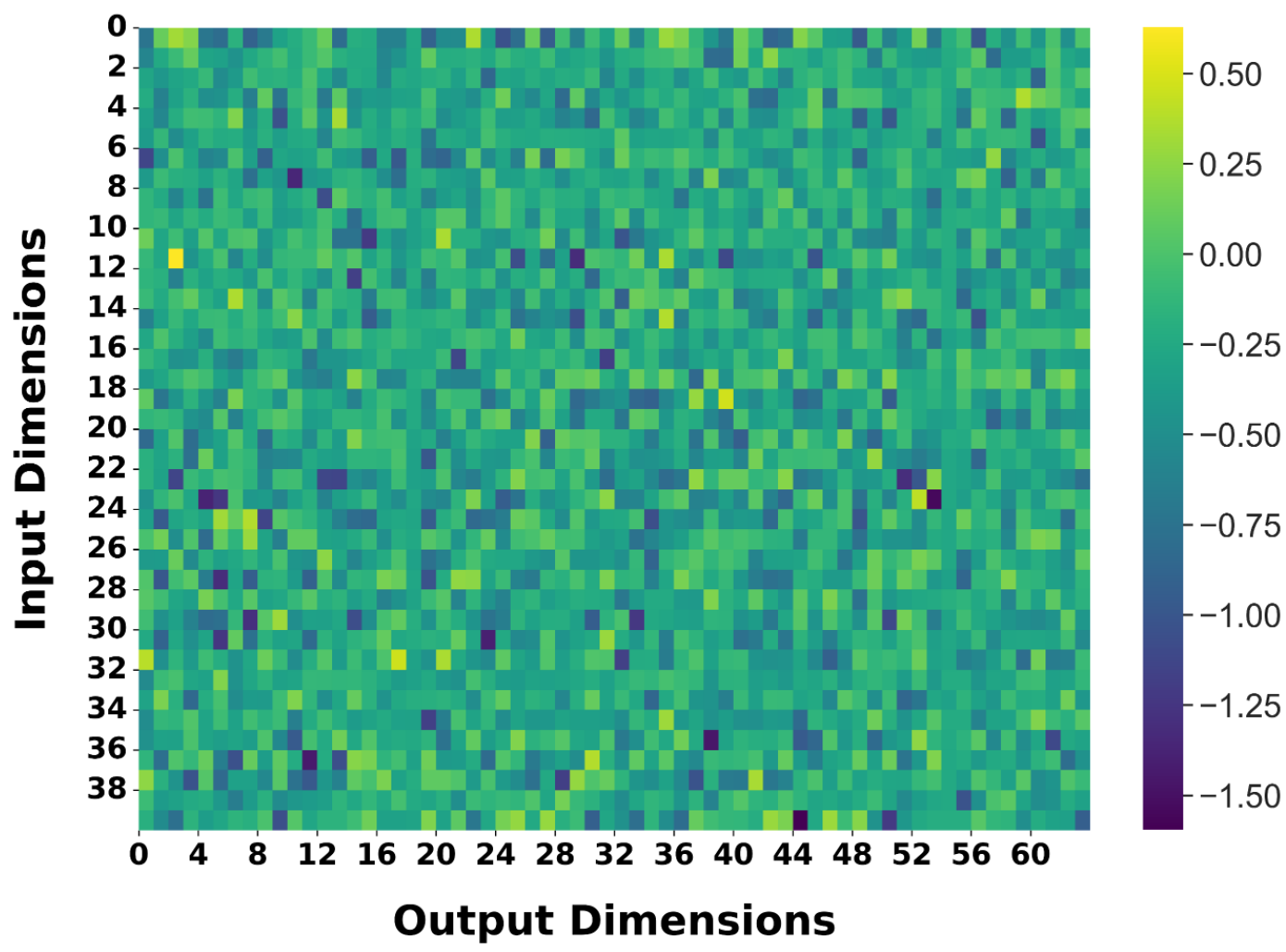

Fig. 10. The weights of readout layers after 200 epochs of 40 x 64 VMM.

### 4.1.1 Effect of the device variation on NVM memristors

For the $TiO_x$-based memristive VMM [2], $X_{LTP}$ and $X_{LTD}$ represent the conductance values for LTP and LTD, respectively, and the following are the equations:

$$X_{LTP} = B\left(1 - e^{\left(-\frac{P}{A}\right)}\right) + X_{\min} \tag{7}$$

$$X_{LTD} = -B\left(1 - e^{\left(\frac{P - P_{\max}}{A}\right)}\right) + X_{\max} \tag{8}$$

$$B = X_{\max} - X_{\min} \Big/ 1 - e^{\left(\frac{-P_{\max}}{A}\right)} \tag{9}$$

The parameters $X_{max}$ and $X_{min}$ denote the maximum and minimum conductance values of the device, which are determined through empirical measurements. $P_{max}$ represents the maximum number of pulses required to transition the device between its extremal conductance states. These parameters are essential for accurately capturing the physical characteristics of the devices in simulations. The parameter $A$ is instrumental in defining the nonlinearity of the weight update mechanism within the VMM system. This parameter can assume both positive and negative values, which are color-coded as black and red, respectively, for illustrative clarity. Figures 2 demonstrate that both LTP and LTD exhibit identical magnitudes but opposite signs for the parameter A, highlighting the symmetric nature of conductance changes in response to positive and negative stimuli. Furthermore, the parameter $B$ is introduced as a function of A, specifically tailored to ensure that the weight update functions conform to the experimentally established bounds of $X_{max}$, $X_{min}$, and $P_{max}$. This relationship underscores the interdependence of these parameters in modelling the complex dynamics of memristive devices within VMM operations [41-42]. By integrating these parameters and equations, the study provides a comprehensive mathematical framework for simulating the behaviour of memristive devices in neuromorphic computing applications. This approach facilitates a deeper understanding of device behaviour under various operational conditions, enabling the development of more efficient and accurate neuromorphic systems.

In the field of NVM memristor devices, the fabrication process often introduces non-ideal characteristics, such as variations in conductance, device-to-device (D2D) inconsistencies, nonlinearity, and programming failures. These imperfections require careful consideration of non-uniform conductance levels and variations during the Neural Networks (NN) simulation. Devices within crossbar arrays are classified according to their conductance profiles to precisely evaluate these non-ideal attributes' impact on training and validation accuracies. The phenomena of nonlinearity and asymmetry in these devices are rigorously defined by equations 7 to 9, where the parameter '$A$' significantly determines the observed nonlinearity in the weight updating process. This inherent nonlinearity, exacerbated by D2D variation, introduces a spectrum of nonlinearity across synaptic devices, complicating NN functionalities' uniform application and predictability.

Particularly within crossbar array configurations, D2D variation poses a substantial challenge. Minor discrepancies in the fabrication process lead to divergent electrical properties among NVM memristor devices, affecting the collective execution of VMM—a foundational operation in NN computations. Such disparities can cause irregular weight updates and unpredictable neural network behaviours. D2D variations undermine the precision of synaptic weights, challenging the achievement of desired computational accuracy. Specifically, networks reliant on accurate timing and weight adjustments, like spiking neural networks, are susceptible to performance declines due to these variations. Moreover, the aggregate effect of D2D variations across a broad crossbar array amplifies these challenges, introducing a significant degree of unpredictability to the system's performance. Illustrations in Figures 11(a) and 11(b), which demonstrate the training and validation accuracies subject to D2D variations ranging from 5% to 30%, underscore the profound influence of device variability. These figures validate the effectiveness of our proposed methods in maintaining exceptional accuracy despite NVM memristor device variations. Figure 11(c) summarizes the training and validation accuracies observed at the concluding phase of the study. Mitigating D2D variation is critical for the advancement of NVM memristor-based computing technologies. It is essential to develop accurate models that reflect these variations and strategies to alleviate their effects. By addressing these challenges, researchers can significantly improve the reliability and efficiency of NVM memristor-based NNs, paving the way for their broader application in computational tasks.

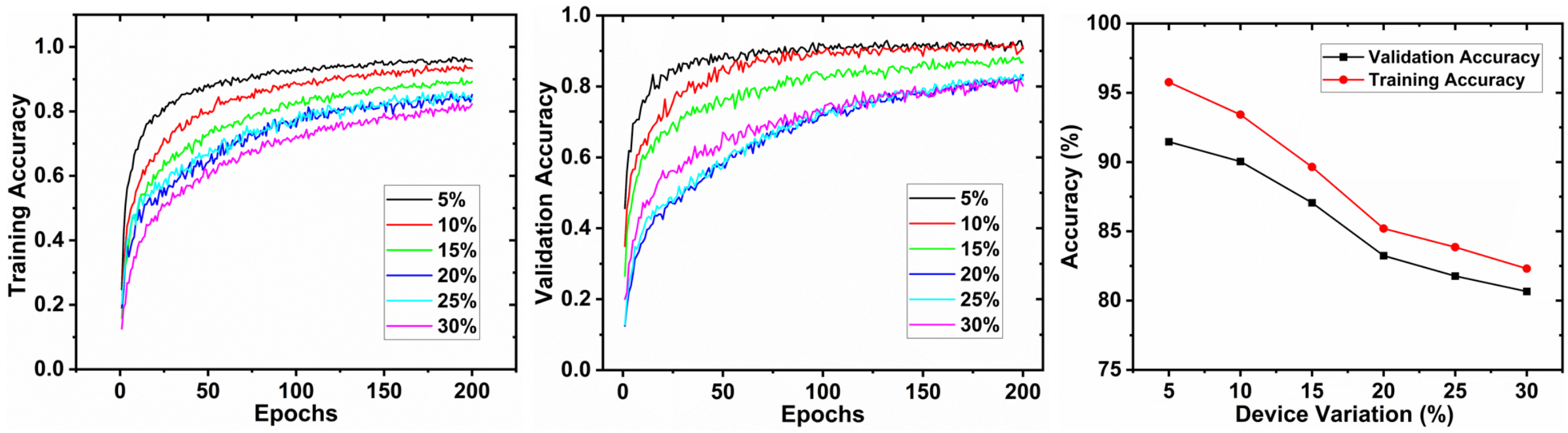

(a) (b) (c)

Fig. 11. (a) Training accuracy (b) validation accuracy per epoch with different D2D levels from 5% to 30 % of VMM (c) Training and validation accuracy decline with increased device variation.

### 4.1.2 Effect of the change in conductance on NVM memristors

The training and validation accuracy for FSDD classification over 200 epochs are presented in Figures 12(a) and 12(b), which depict the relationship between the number of epochs and accuracy for three distinct conductance regions in $TiO_x$ memristor devices within crossbar arrays [41]. These figures provide insight into the impact of conductance variability on the performance of RC systems. From the figures, the delineated regions correspond to different ranges of conductance change that can arise from device-to-device variability, fabrication imperfections, or programming inconsistencies. Figures 12(a) and 12(b) show curves of training accuracy of the FSDD experiment under three regions with different approximate ranges. Region 1, Region 2, and Region 3 represent approximately regions with conductance ranges [-1.30 × $10^{-5}$ S, 0.40 × $10^{-5}$ S], [-1.20 × $10^{-5}$ S, 0.3 × $10^{-5}$ S], and [-1.10 × $10^{-5}$ S, 0.30 × $10^{-5}$ S], respectively.

Region 1 demonstrates the highest training and validation accuracies, suggesting that a broader conductance range affords a more dynamic and flexible synaptic weight adaptation during the learning process. This adaptability could facilitate a more robust feature representation within the RC system, enhancing classification accuracy. In contrast, regions 2 and 3 exhibit moderate training and validation accuracy levels. The more restricted conductance range, relative to region 1, implies a potential limitation in the network's learning capability, which may manifest as reduced accuracies. The comparative analysis of these regions underscores the importance of conductance range and distribution within memristor devices, particularly in crossbar arrays, as critical factors affecting NN learning dynamics and generalization ability. Wider conductance ranges benefit NN performance, likely due to the greater flexibility in synaptic weight scaling they offer during the training phase.

Conclusively, variations in conductance within $TiO_x$ memristors directly influence the effectiveness of weight updates within crossbar memristor arrays, subsequently affecting the learning proficiency and accuracy of the implemented NN. Optimizing conductance ranges and ensuring consistency across the array is imperative for enhancing NN performance in $TiO_x$ memristor technology applications.

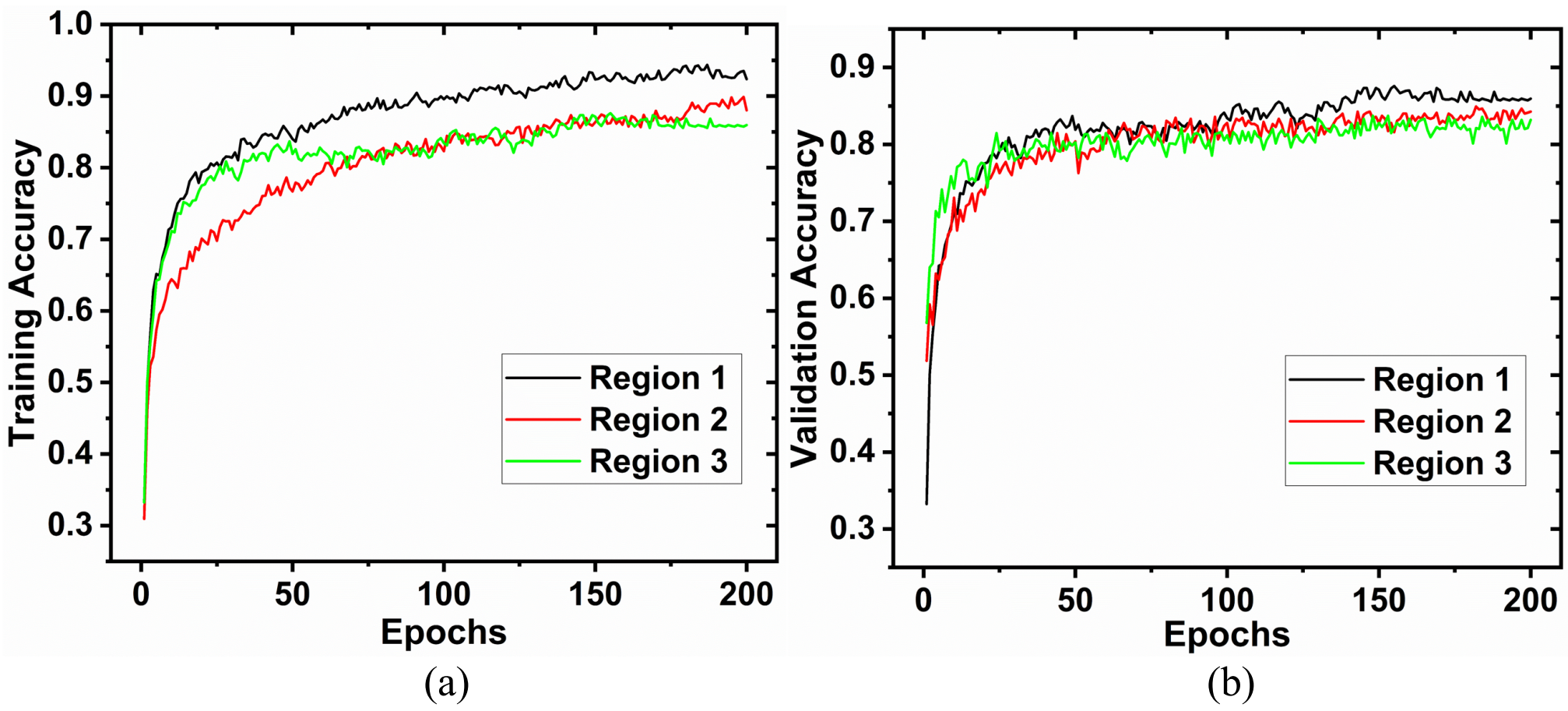

Fig. 12. (a) Training accuracy (b) validation accuracy per epoch across different regions with varying conductance ranges.

## 4.2 Mackey-glass Series Forecasting

The MG time series is a foundational benchmark in time series prediction and analysis, widely utilized for assessing the predictive performance of diverse computational models, such as neural networks and machine learning algorithms. Introduced by Michael Mackey and Leon Glass in 1977, the MG time series is derived from a differential equation to simulate the regulation of white blood cells in the bloodstream [30]. It is characterized by a non-linear time-delay differential equation, embodying intricate dynamics that pose significant predictive challenges. This complexity makes the MG time series a critical tool for evaluating the efficacy of forecasting methodologies in capturing and anticipating complex temporal patterns.

The equation is defined as follows [44]:

$$\frac{dx(t)}{dt} = \beta \frac{x(t-\tau)}{1+x(t-\tau)^n} - \gamma x(t) \tag{10}$$

where *x(t)* represents the quantity of interest at time *t*, $\beta$ and $\gamma$ are positive constants, $\tau$ is the delay parameter, and *n* is a nonlinearity parameter. The equation models the rate of change of *x(t)* over time as a function of its own past values, introducing a delay ($\tau$) that makes the system's future behaviour dependent on its past states. This delay, combined with the nonlinearity introduced by the term $x(t-\tau)^n$ in the denominator, produces various dynamical behaviours, including chaotic regimes, making the MG time series a compelling subject for predictive modelling.

In our framework, we have chosen $\sigma = 0.2$, $\beta = 0.1$, $N = 10$, $\tau = 17$ and $x(0) = 1.2$. The implemented RC system, incorporating 10 DM elements, offers a sophisticated method for capturing and processing the temporal dynamics inherent in the MG dataset [45]. The DM, functioning within the reservoir, encodes the input time series into a high-dimensional state space. Within the reservoir, the DM components are responsible for transforming the input time series into a high-dimensional state space, leveraging the memristive devices' non-linear and state-dependent conductance characteristics. This transformation effectively captures and preserves the temporal information present in the MG series, reflecting the series' historical data points and nuanced fluctuations over time. A notable advancement in our approach is the integration of a $TiO_x$ memristor-based VMM accelerator as the readout layer. The readout leverages the memristor's synaptic-like properties, renowned for their biomimetic attributes, to perform an efficient linear conversion of the reservoir states into subsequent output predictions [46]. Figure 13(a) displays the convergence of training and validation loss throughout the epochs for the MG time series prediction model. Figure 13(b) illustrates the model's training against the actual MG series curve, learning to anticipate the curve's progression utilizing the established parameter values and the RC framework. Finally, Figure 13(c) showcases the trained model's predictions, endeavouring to mirror the target curve accurately, thereby demonstrating the model's predictive prowess within the specified system parameters. The system maintained a low normalized root mean square error (NRMSE) of 0.036 in time series prediction, showcasing its capability.

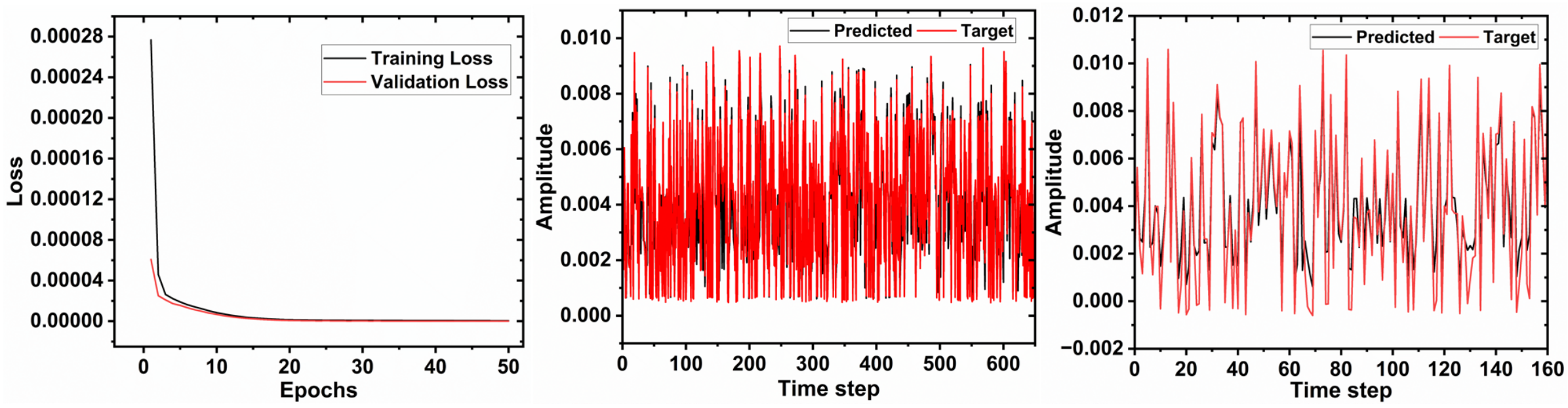

(a) (b) (c)

Fig. 13. (a) Convergence of training and validation loss over epochs. (b) Improvement in model training to enhance predictive accuracy (c) Predictive dynamics of the model for MG forecasting at $\tau = 17$.

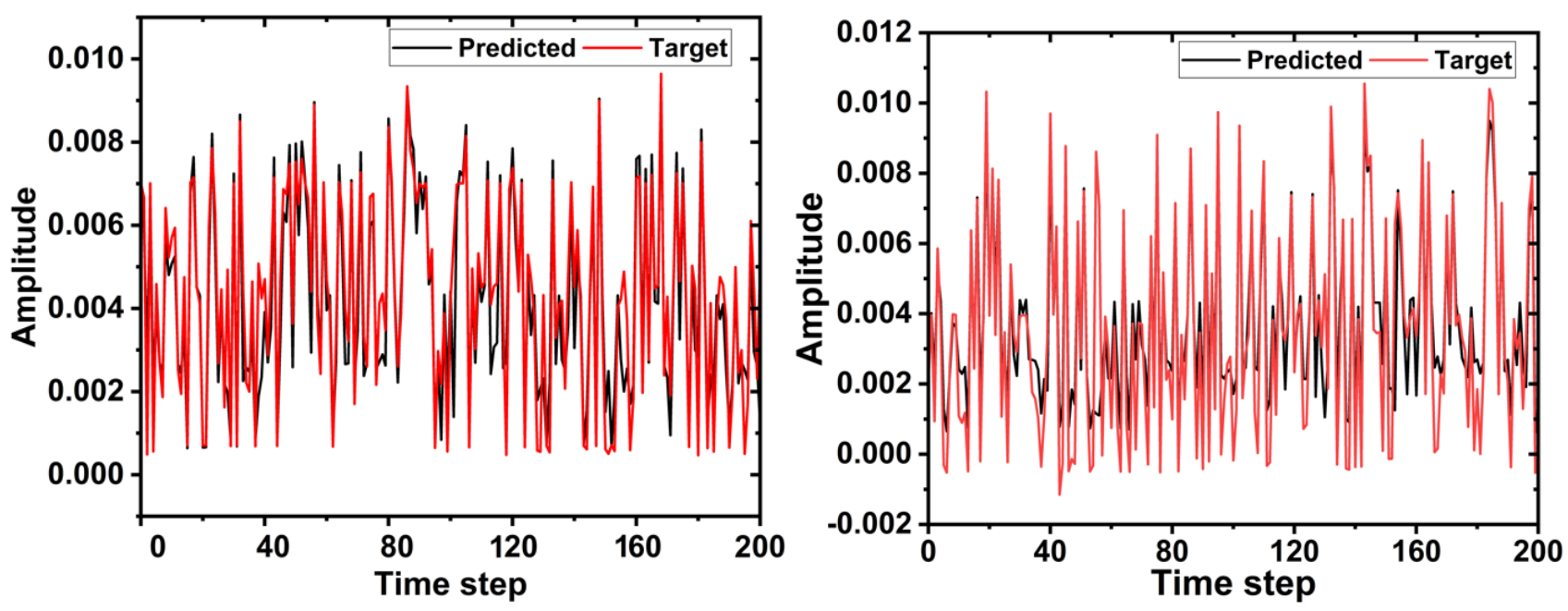

(a) (b)

Fig. 14. Predictive dynamics of the model for MG forecasting at (a) $\tau = 12$ (b) $\tau = 15$.

The MG system task is formulated as a one-step prediction problem. In addition, we have expanded our evaluation by including prediction tasks with varying time delays ($\tau$) to provide a more comprehensive assessment of the model's memory performance. Specifically, we have added results for τ=12, and τ=15, as illustrated in Fig. 14(a) and (b), respectively. This extended analysis demonstrates the impact of different time delays on prediction accuracy and further underscores the long-term memory capabilities of the $TiO_x$-based memristor in our design. Figure 15 presents a comparative visualization of chaotic dynamics through two adjacent scatter plots: one representing the actual behaviour of a time series as governed by the MG differential equation and the other depicting the predicted behaviour as determined by an RC model. The left scatter plot (marked with blue points) illustrates the genuine chaotic dynamics of the MG time series. Each point in this plot corresponds to a pair of values, $y(t-\tau)$ and $y(t)$, derived from the authentic time series data [47]. Here, $\tau$ signifies the time delay parameter, a defining feature of the MG system. This plot serves as a benchmark, capturing the intrinsic chaotic behaviour of the system. On the right, the scatter plot (indicated by red points) conveys the chaotic behaviour as forecasted by the RC model. Mirroring the structure of the left plot, this graph showcases a scatter plot of the model's predicted values, with each point representing a pair: the predicted $y(t-\tau)$ and $y(t)$. The alignment of this plot with the left plot demonstrates the RC model's proficiency in emulating the complex, chaotic dynamics characteristic of the MG system, leveraging learned patterns from the training dataset. These plots visually assess the RC model's capability to replicate chaotic time series dynamics, an essential aspect of modelling real-world phenomena that exhibit chaotic behaviour.

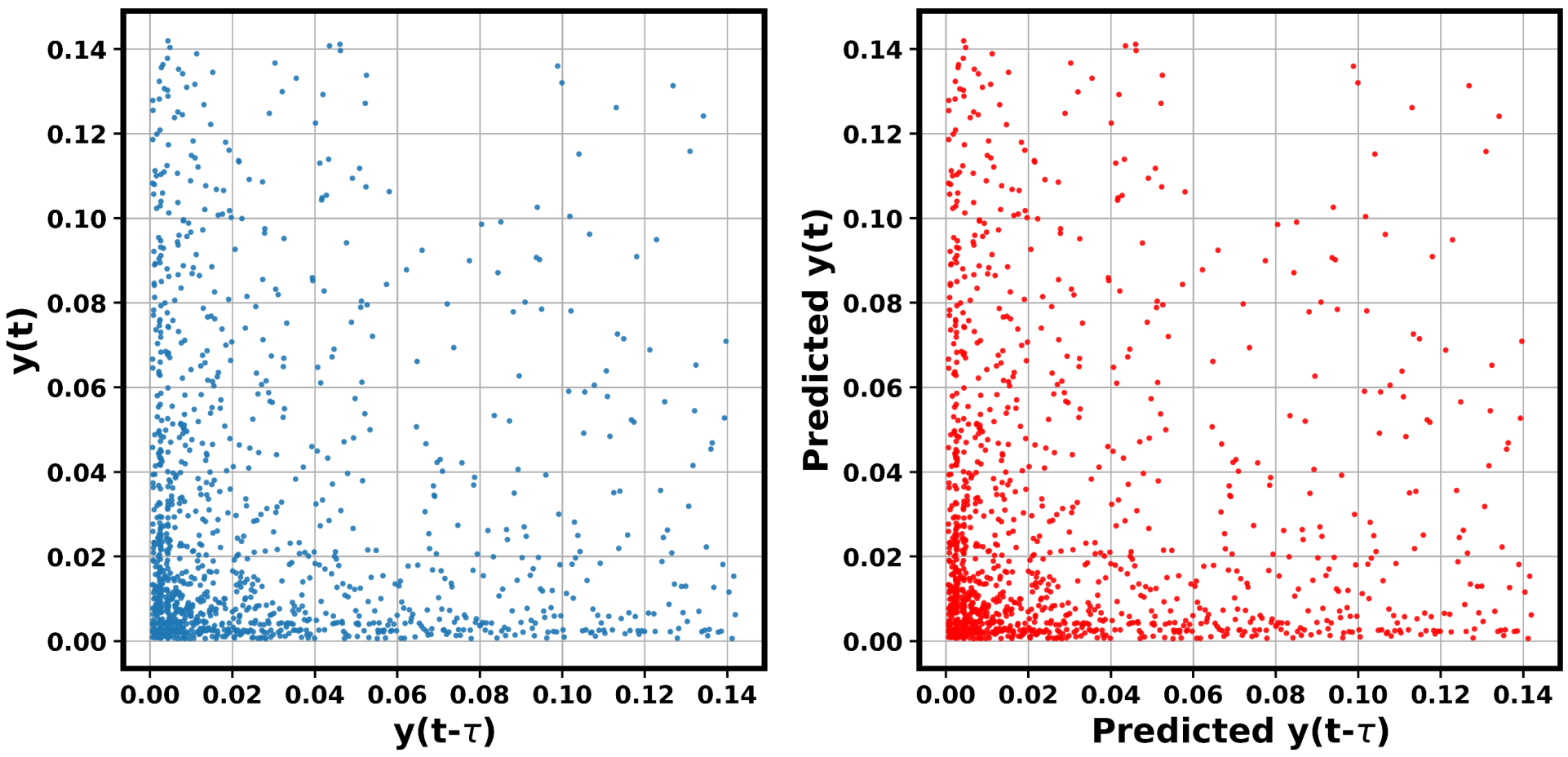

(a) (b)

Fig. 15. Comparative analysis of (a) actual (b) predicted chaotic dynamics of the MG time series.

Table 1: Comparison of proposed RC with earlier design.

| Ref. | Devices material | No. of memristor | Time constant | Ron (Ω) | Device variation | Tasks | Recognition ratio |
|---|---|---|---|---|---|---|---|
| **30** | DM: $WO_x$ | 50 | 50 ms | -- | -- | Spoken-digit recognition | 99.2% |
| **49** | DM: $WO_x$ | DM: 88<br>NVM: 176 x 10 | 50 ms | DM: $10^6$ | -- | 1.Handwritten digit recognition<br>2. Nonlinear dynamic task | 1. 88.1%<br>2. NRMSE: 0.031 |
| **50** | DM: $SiO_2$<br>NVM: Ta2O5 | DM: 110<br>NVM: 110 X 10 | 3 ms | DM: $10^4$<br>NVM: $10^6$ | -- | Handwritten digit recognition | 83 % |
| **51** | SnS | 5 | >10 ms | DM: $10^6$ | -- | Korean sentence recognition | 91 % |
| **52** | DM: BFO<br>NVM: SRO | DM: 5<br>NVM: 6 X 10 | 2 ms | DM: $10^6$<br>NVM: $10^7$ | 56% | 1. MNIST Handwritten digit recognition<br>2. Hénon map time-series prediction | 1. 89.5 %<br>2. NRMSE: 0.017 |
| **53** | DM: Pt/Co | DM: 1 | < 5 ns | -- | -- | Handwritten digit recognition | 87.6 % |
| **This Work** | DM: $WO_x$<br>NVM: TiOx | DM: 40<br>NVM: 40 x 64 | 50 ms | DM: 2 x $10^3$<br>NVM: $10^4$ | 30% | 1. Speech digit recognition<br>2. Mackey-glass forecasting | 1. 98.84 %<br>2. NRMSE: 0.036 |

DM: Dynamic memristor; NVM: Non-volatile memristor;

## 5. Comparison and Discussion

In this work, we have performed a detailed evaluation of accuracy, device utilization, and robustness against device variations for multiple tasks. Table 1 presents a comparative analysis of our framework against previous implementations. Our approach leverages a compact configuration of 40 DM and a 40 x 64 NVM matrix, achieving a 98.84% recognition ratio for speech digit recognition and a NRMSE of 0.036 for Mackey-Glass time-series forecasting. In comparison, previous works employed larger arrays (110 DM and 110x10 NVM in [50]) but achieved significantly lower performance metrics, such as an 83% recognition ratio for handwritten digit recognition. Moreover, our design incorporates real-world fabrication non-idealities by considering a 30% variation in device parameters, which is crucial for practical deployment. The proposed framework demonstrates a power-efficient design with a Ron value of $2\times10^3$ Ω and $10^4$ Ω, which is optimized compared to other configurations like [49] (Ron of $10^6$ Ω) and [52] (Ron of $10^7$ Ω). This reduction in Ron helps minimize power consumption in large-scale arrays. In summary, our framework demonstrates with fewer devices and lower power consumption, making it a strong candidate for resource-efficient neuromorphic computing applications.

## 6. Conclusion

In this paper, we introduce a comprehensive analog memristive reservoir computing (RC) framework characterized by an integrated input module, a reservoir module, and a readout module. The innovative architecture of the proposed system primarily leverages the concept of a reservoir to enrich the state representation significantly through multiple parallel masking processes. This is further bolstered by the RC structure, which underpins the system's substantial memory retention capabilities. Central to our architecture's functionality are two distinct types of memristors: the $WO_x$-based dynamic memristor (DM) and the $TiO_x$-based non-volatile memories (NVM), each fulfilling a critical role within the circuit framework. The DM are instrumental for their complex dynamic behaviours—facilitated by diffusion effects—that act as reservoir nodes, thereby projecting the original input signals into an expansive high-dimensional state space. Conversely, the NVM are employed in crafting a memristive crossbar array, which adeptly facilitates the weighted summation of reservoir states, enhancing the computational efficiency of the system. A distinguishing feature of the proposed system is its operation entirely in the analog domain, eschewing digital conversions to maintain signal integrity and processing continuity. The efficacy and robustness of the proposed RC system are demonstrated through its application to two benchmark tasks: the recognition of isolated spoken digits with partial inputs and the prediction of Mackey-Glass time series. The system not only achieved a remarkable training accuracy of 98.84% for spoken-digit recognition but also maintained a low normalized root mean square error (NRMSE) of 0.036 in time series prediction, showcasing its proficiency. These compelling results underscore the significant potential of fully analog memristor-based RC systems in efficiently handling intricate temporal processing tasks. This research lays a solid groundwork for future explorations and advancements in analog computing architectures, marking a step forward in developing sophisticated temporal pattern recognition and prediction systems.

## 7. Acknowledgements

This work was supported in part by the Nano-Material Technology Development Program through the National Research Foundation of Korea (NRF) funded by the Ministry of Science and ICT under Grant NRF2022M3H4A1A01009658 and by the National Research Foundation (NRF) funded by the Korean government (MSIT) (No. RS-2024-00439520).